\documentclass[sigconf,screen]{acmart}
\AtBeginDocument{%
  }
\usepackage{tabularx}
\setcopyright{none}
\renewcommand\footnotetextcopyrightpermission[1]{}
\acmConference[MM '26]{The 34th ACM International Conference on Multimedia}{November 10--14,
  2026}{Rio de Janeiro, Brazil}

\begin{document}

\title{TowerDataset: A Heterogeneous Benchmark for Transmission Corridor Segmentation with a Global-Local
  Fusion Framework}


\author{Xu Cui}
\email{cuixu18@stu.hit.edu.cn}
\orcid{0009-0007-2083-2188}
\affiliation{%
  \institution{Harbin Institute of Technology (Weihai)}
  \city{Weihai}
  \country{China}
}

\author{Xinyan Liu}
\orcid{0000-0003-2638-4324}
\email{xinyliu@hit.edu.cn}
\affiliation{%
  \institution{Harbin Institute of Technology (Weihai)}
  \city{Weihai}
  \country{China}
}

\author{Chen Yang}
\orcid{0009-0000-4052-3795}
\email{yangchen212@mails.ucas.ac.cn}
\affiliation{%
  \institution{University of the Chinese Academy of Sciences}
  \city{Beijing}
  \country{China}
}

\author{Zhaobo Qi}
\orcid{0000-0001-9196-9818}
\email{qizb@hit.edu.cn}
\affiliation{%
  \institution{Harbin Institute of Technology (Weihai)}
  \city{Weihai}
  \country{China}
}

\author{Beichen Zang}
\orcid{0000-0001-5030-0632}
\email{beiczhang@hit.edu.cn}
\affiliation{%
  \institution{Harbin Institute of Technology (Weihai)}
  \city{Weihai}
  \country{China}
}

\author{Weigang Zhang}
\email{wgzhang@hit.edu.cn}
\orcid{0000-0003-0042-7074}
\affiliation{%
  \institution{Harbin Institute of Technology (Weihai)}
  \city{Weihai}
  \country{China}
}

\author{Antoni B. Chan}
\email{abchan@cityu.edu.hk}
\orcid{0000-0002-2886-2513}
\affiliation{%
  \institution{City University of Hong Kong}
  \city{Hong Kong}
  \country{China}
}




\renewcommand{\shortauthors}{Cui et al.}

\begin{abstract}
Fine-grained semantic segmentation of transmission-corridor point clouds is fundamental for intelligent power-line inspection. However, current progress is limited by realistic data scarcity and the difficulty of modeling global corridor structure and local geometric details in long, heterogeneous scenes. Existing public datasets usually provide only a few coarse categories or short cropped scenes which overlook long-range structural dependencies, severe long-tail distributions, and subtle distinctions among safety-critical components. As a result, current methods are difficult to evaluate under realistic inspection settings, and their ability to preserve and integrate complementary global and local cues remains unclear. 
To address above challenges, we introduce \textbf{TowerDataset}, a heterogeneous benchmark for transmission-corridor segmentation. TowerDataset contains $661$ real-world scenes and about $2.466$ billion points. It preserves long corridor extents, defines a fine-grained $22$-class taxonomy, and provides standardized splits and evaluation protocols. 
In addition, we present a \textbf{global-local fusion framework} which preserves and fuses whole-scene and local-detail information. A whole-scene branch with NoCrop training and prototypical contrastive learning captures long-range topology and contextual dependencies. A block-wise local branch retains fine geometric structures. Both predictions are then fused and refined by geometric validation. This design allows the model to exploit both global relationships and local shape details when recognizing rare and confusing components. Experiments on TowerDataset and two public benchmarks demonstrate the challenge of proposed benchmark and the robustness of our framework in real, complex, and heterogeneous transmission-corridor scenes. The dataset will be released soon at: \url{https://huggingface.co/datasets/tccx18/Towerdataset/tree/main}.
\end{abstract}

\ccsdesc[500]{Computing methodologies~Image segmentation}

\keywords{point cloud, power-line inspection, fine-grained semantic segmentation, global-local fusion}

\maketitle

\begin{figure}[t]
  \centering
  \includegraphics[width=1\columnwidth]{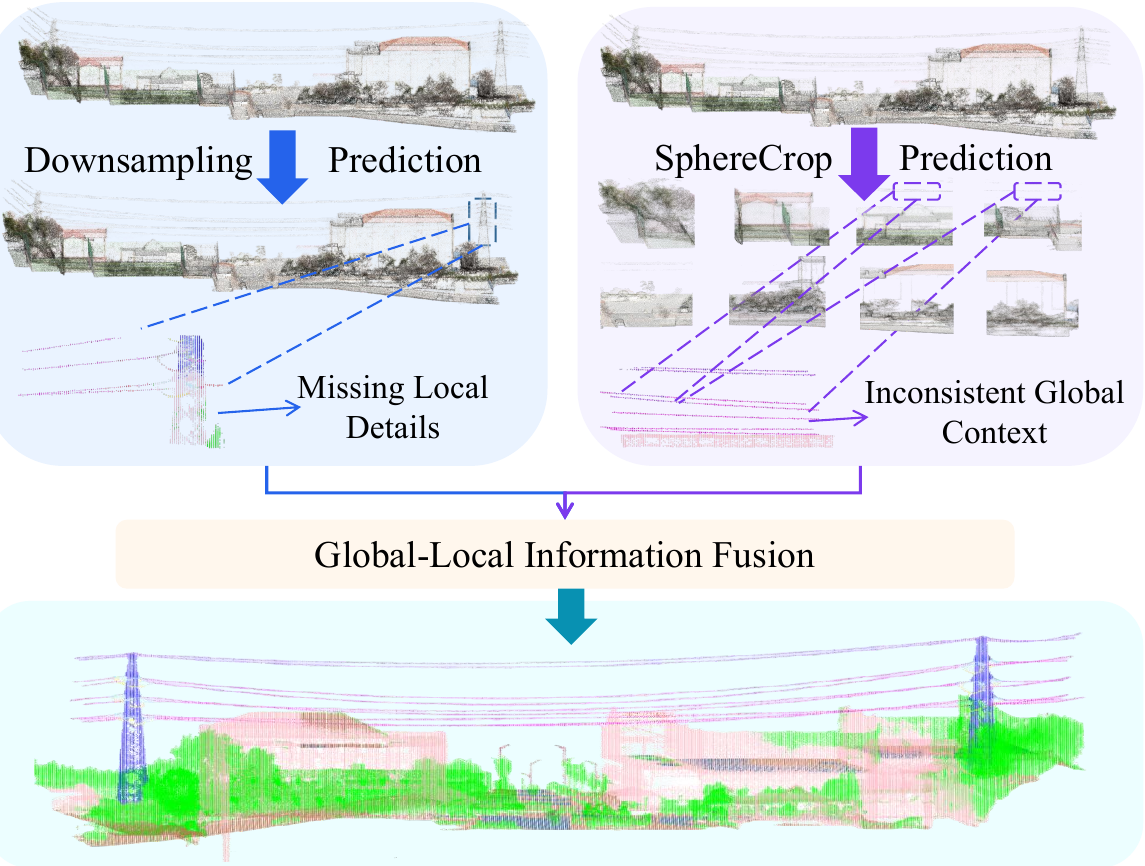}
  \caption{Relying only on local processing can break long-range structural continuity and weaken global contextual consistency, while relying only on global reasoning can miss fine-grained local geometric details}
  \Description{A motivation figure showing that local processing may lose global context and global reasoning may weaken local geometric detail, motivating global-local fusion for fine-grained transmission-corridor segmentation}
  \label{fig:motivation}
\end{figure}

\section{Introduction}
\label{sec:intro}


Semantic segmentation of transmission-corridor point clouds is essential for intelligent power-line inspection, supporting vegetation-risk warning, tower-condition assessment, asset inventory, and maintenance planning. Despite rapid progress in 3D scene understanding on benchmarks such as ScanNet~\cite{scannetv2}, SemanticKITTI~\cite{semantickitti}, and SensatUrban~\cite{sensaturban}, with advances in unified segmentation, sparse modeling, boundary reasoning, and parameter-efficient transfer, applying these methods to power-infrastructure scenes~\cite{eclair,ts40k,gridnethd} remains difficult. Transmission corridors involve long spatial extents, thin elongated structures, compact fittings, cluttered natural backgrounds, and substantial geometric variation.


One limitation is the lack of realistic benchmarks. Existing power-line datasets usually provide only a few coarse categories or short cropped scenes~\cite{powerline_ehvtl,ca_pointnetpp,emafl_ptv3}, which support coarse asset extraction but do not fully reflect practical inspection needs. In real corridors, conductors, jumpers, optical cables, and different insulator types may look locally similar while serving different functions. Cropped scenes also weaken long-range dependencies and underrepresent the long-tail distributions common in corridor data~\cite{focalloss,classbalanced,balancedsoftmax}. As a result, current methods are not adequately evaluated under realistic fine-grained settings.

Another limitation lies in modeling. Transmission-corridor segmentation must preserve the continuity of large cross-scene objects while remaining sensitive to sparse and small components. This requires global context for attachment relationships, height patterns, and corridor topology, as well as local geometric cues for small-object recognition and fine-grained discrimination. As shown in Fig.~\ref{fig:motivation}, many pipelines still rely on block-wise partitioning for memory control~\cite{kpconv,randlanet,minkowski,ptv3,ca_pointnetpp,bdfnet}, which can fragment long wires and discard scene-level context. Whole-scene reasoning alone, however, may weaken local detail. The challenge is further amplified by severe class imbalance, with dominant background classes covering most points and safety-critical components remaining sparse.

To address these limitations, we contribute on both the data and model sides. We first introduce \textbf{TowerDataset}, a heterogeneous benchmark with $661$ real-world scenes and approximately $2.466$ billion points. It preserves long corridor extents, defines a fine-grained $22$-class taxonomy, and provides standardized splits and evaluation protocols. We then propose a \textbf{global-local fusion framework}. A NoCrop whole-scene branch with prototypical contrastive learning captures long-range context, and a block-wise local branch preserves fine geometry. Their predictions are fused and refined by geometric validation, enabling joint use of structural context and local shape evidence for rare and confusing components.

Experiments on TowerDataset and two public benchmarks~\cite{eclair,toronto3d} show that TowerDataset offers a challenging fine-grained evaluation setting and that the proposed framework performs effectively across different corridor scenes. The main contributions are as follows:

\textbf{1) A heterogeneous fine-grained benchmark.} We establish TowerDataset, a large-scale transmission-corridor benchmark with $661$ scenes, $2.466$ billion points, a fine-grained $22$-class semantic taxonomy, and standardized evaluation protocols for realistic component-level research.

\textbf{2) A global-local fusion framework.} We propose a framework that combines NoCrop whole-scene learning, prototypical contrastive supervision, block-wise local refinement, and geometric validation to preserve both structural continuity and local discriminability.

\textbf{3) Extensive evaluation in realistic scenes.} We provide extensive experiments on TowerDataset and public benchmarks to show both the challenge of TowerDataset and the effectiveness of the proposed framework.




\section{Related Work}
\label{sec:related}

\subsection{Transmission-Corridor Segmentation and Datasets}
Recent transmission-corridor research has shifted from coarse target extraction to deep models for component-level understanding. CA-PointNet++~\cite{ca_pointnetpp} introduces coordinate attention, BDF-Net~\cite{bdfnet} strengthens discrimination with bilinear distance features, and EMAFL-PTv3~\cite{emafl_ptv3} combines multi-scale attention with focal supervision for key-category segmentation. More recent studies further explore scene-specific designs, including foreground separation, key-region modeling along transmission lines, and graph-attention encoding for fine-grained corridor LiDAR data~\cite{filterassist_tl,tlsnet,gkcae}. In parallel, public benchmarks such as ECLAIR, TS40K, and GridNet-HD~\cite{eclair,ts40k,gridnethd} broaden evaluation from aerial corridor scenes to rural and multimodal infrastructure perception.
However, most existing methods still emphasize limited coarse classes or isolated designs, leaving fine-grained recognition and structural continuity insufficiently addressed. Our work targets this gap with a unified framework for long-span scenes and fine-grained labels.

\subsection{3D Semantic Segmentation Backbones}
General 3D semantic segmentation has progressed from hierarchical point learning and sparse convolution to transformer-based and structure-aware models. PointNet++~\cite{pointnetpp} established hierarchical local aggregation on raw points, and MinkowskiNet~\cite{minkowski} demonstrated the scalability of sparse voxel convolution. Transformer-based backbones such as PTv3~\cite{ptv3}, OneFormer3D~\cite{oneformer3d}, and ODIN~\cite{odin} further strengthen large-scene representation through contextual modeling. Other studies explore adaptive sparse convolution and structural priors~\cite{oacnns,curvecloudnet,bfanet}, while efficient LiDAR segmentation and robust optimization are investigated in Campoint, CDSegNet, and RangeRet~\cite{campoint,cdsegnet,rangeret2026}. Unified research frameworks such as Pointcept~\cite{pointcept} further make these families easier to compare under shared implementations. 

However, most of these pipelines still depend on block-wise partitioning for memory control, which can break the structural continuity of kilometer-scale corridors. Recent pretraining methods such as Point-MAE~\cite{pointmae}, MaskClu~\cite{maskclu2026}, and MSC~\cite{msc} improve transferable 3D representations. Adaptation-oriented studies also investigate GFT~\cite{gft2026}, PonderV2~\cite{ponderv2}, PMA~\cite{pma}, and GCDLSS~\cite{kim2026gcdlss}, but they do not resolve the scene-level topology and category ambiguity in industrial settings. In contrast, our method preserves global corridor continuity through a NoCrop whole-scene branch with whole-scene downsampling, while a local-detail branch and prediction fusion recover fine structures and improve discrimination of confusing attachment classes.

\subsection{Long-Tail and Contrastive Learning}
Long-tail learning is critical in corridor scenes where dominant classes like ground and vegetation overwhelm sparse but safety-critical components. Focal Loss~\cite{focalloss}, LDAM~\cite{ldam}, and Balanced Meta-Softmax~\cite{balancedsoftmax} address class imbalance, while Prototypical Networks~\cite{protonet}, SupCon~\cite{supcon}, and recent language-guided point-cloud segmentation methods~\cite{epsegfz2026} improve feature discrimination through class-centered embeddings and contrastive separation.
However, these methods are usually applied as local-sample objectives rather than coupled with whole-scene corridor modeling, limiting their ability to distinguish components that share similar local geometry but play different roles in global topology. In contrast, our method integrates rare-class reweighting, focal reweighting, and prototype-based contrastive learning into a NoCrop whole-scene branch, so sparse attachment classes are learned with stronger supervision and scene-level contextual cues under extreme imbalance.

\begin{table*}[t]
\centering
\caption{Comparison of TowerDataset with other transmission-line related datasets}
\label{tab:dataset_comparison}
\small
\setlength{\tabcolsep}{4pt}
\begin{tabular*}{\textwidth}{@{\extracolsep{\fill}}p{0.15\textwidth} p{0.12\textwidth} p{0.56\textwidth} p{0.07\textwidth}@{}}
\toprule
Dataset & Modality & Scene and scale & Classes \\
\midrule
TowerDataset & 3D point cloud & $661$ real transmission-corridor scenes, $2.466$B points, long real spans & $22$ \\
ECLAIR~\cite{eclair} & Aerial LiDAR & Aerial power-line infrastructure scenes, $10$ km$^2$, about $582$M points & $11$ \\
TS40K~\cite{ts40k} & 3D point cloud & Rural terrain and electrical transmission-system samples, $24{,}355$ samples, $2.595$B points & $22$ \\
GridNet-HD~\cite{gridnethd} & LiDAR + images & Multimodal power-line infrastructure zones, $36$ zones, $7{,}694$ images, $2.449$B points & $11$ \\
EHVTL~\cite{powerline_ehvtl} & 3D point cloud & Extra-high-voltage transmission-line corridor scenes & $6$ \\
EMAFL-PTv3~\cite{emafl_ptv3} & 3D point cloud & Transmission-line spans, $40$ spans & $5$ \\
\bottomrule
\end{tabular*}
\end{table*}

\section{The TowerDataset Benchmark}
\label{sec:tower_dataset}

To facilitate reliable evaluation of fine-grained transmission-corridor segmentation, we introduce \textbf{TowerDataset}, a large-scale point cloud benchmark specifically for intelligent power-line inspection. Unlike existing datasets that prioritize coarse-grained classification of dominant terrain features, TowerDataset provides finer component-level annotations, realistic long-tail category distributions, and standardized evaluation splits for power-corridor scenes.

TowerDataset includes $661$ scenes with a total of about $2.466$ billion points. These scenes are divided into $419$ for training, $122$ for validation, and $120$ for testing. Many other datasets cut power lines into small tiles of $50$ to $100$ meters. In contrast, TowerDataset keeps real-world corridor scenes. The middle length of these scenes is $355.8$ meters, while the 90th percentile length is $641.5$ meters and the maximum length is $992.1$ meters. The number of points in each scene varies from $0.31$ million to $61.57$ million. The density of points ranges from $11.08$ to $2306.33$ points per square meter, and the middle density value is $93.78$ points per square meter.

\subsection{Annotation Categories}

TowerDataset defines a comprehensive \textbf{22-class semantic categories}. Unlike many current datasets that merge features into a handful of primary classes, ours maintains the fine details of both power infrastructure and its surroundings which makes it more suitable for practical research.

As shown in Table~\ref{tab:dataset_comparison}, TowerDataset offers several advantages over existing transmission-line related datasets. Compared with EHVTL and EMAFL-PTv3, which focus on $5$--$6$ classes, TowerDataset provides a richer fine-grained label space with $22$ categories. Compared with TS40K, which is large but mainly sample-based, TowerDataset preserves long real corridor spans and therefore retains the structural continuity for topology-aware reasoning. Compared with GridNet-HD, which emphasizes multimodal infrastructure perception, TowerDataset focuses on fine-grained component-level segmentation in pure point-cloud corridor scenes. Compared with ECLAIR, TowerDataset covers a broader set of corridor components and stronger engineering heterogeneity, including multiple voltage levels, diverse tower types, and complex real-scene conditions. These characteristics make TowerDataset a more challenging and realistic benchmark for evaluating long-range structure preservation, rare-component recognition, and practical transmission-corridor understanding.

These advantages are also reflected in the category statistics of real corridor scenes. The $22$ categories in TowerDataset exhibit an extreme long-tail distribution consistent with practical transmission-corridor data. Background categories dominate the point population, accounting for more than $94\%$ of all points, while several fine-grained attachment classes together occupy only about $0.16\%$. This severe imbalance makes TowerDataset a challenging benchmark for long-tail learning and fine-grained component recognition. The fine-grained labels also support both comprehensive full-scene evaluation and focused core-infrastructure assessment, providing flexibility for different research objectives.

\section{Method}
\label{sec:method}

\subsection{Problem Definition}

Given a point cloud scene $\mathcal{P}=\{(\mathbf{x}_i, y_i)\}_{i=1}^N$, where $\mathbf{x}_i\in\mathbb{R}^3$ denotes the XYZ coordinate of point $i$ and $y_i\in\{0,\dots,C-1\}$ is the corresponding label, let $X=\{\mathbf{x}_i\}_{i=1}^N$ denote the input point set. Fine-grained semantic segmentation aims to predict a label $\hat{y}_i$ for each point:

\begin{equation}
\hat{y}_i = \arg \max_{c} p(y_i=c \mid X, \Theta), \quad c \in \{0,\dots,C-1\},
\end{equation}
where $p(y_i=c \mid X, \Theta)$ is the probability that the segmentation model assigns class $c$ to point $\mathbf{x}_i$ conditioned on the whole input scene $X$ under parameters $\Theta$.

\subsection{Framework Overview}

\begin{figure*}[t]
  \centering
  \includegraphics[width=1\textwidth]{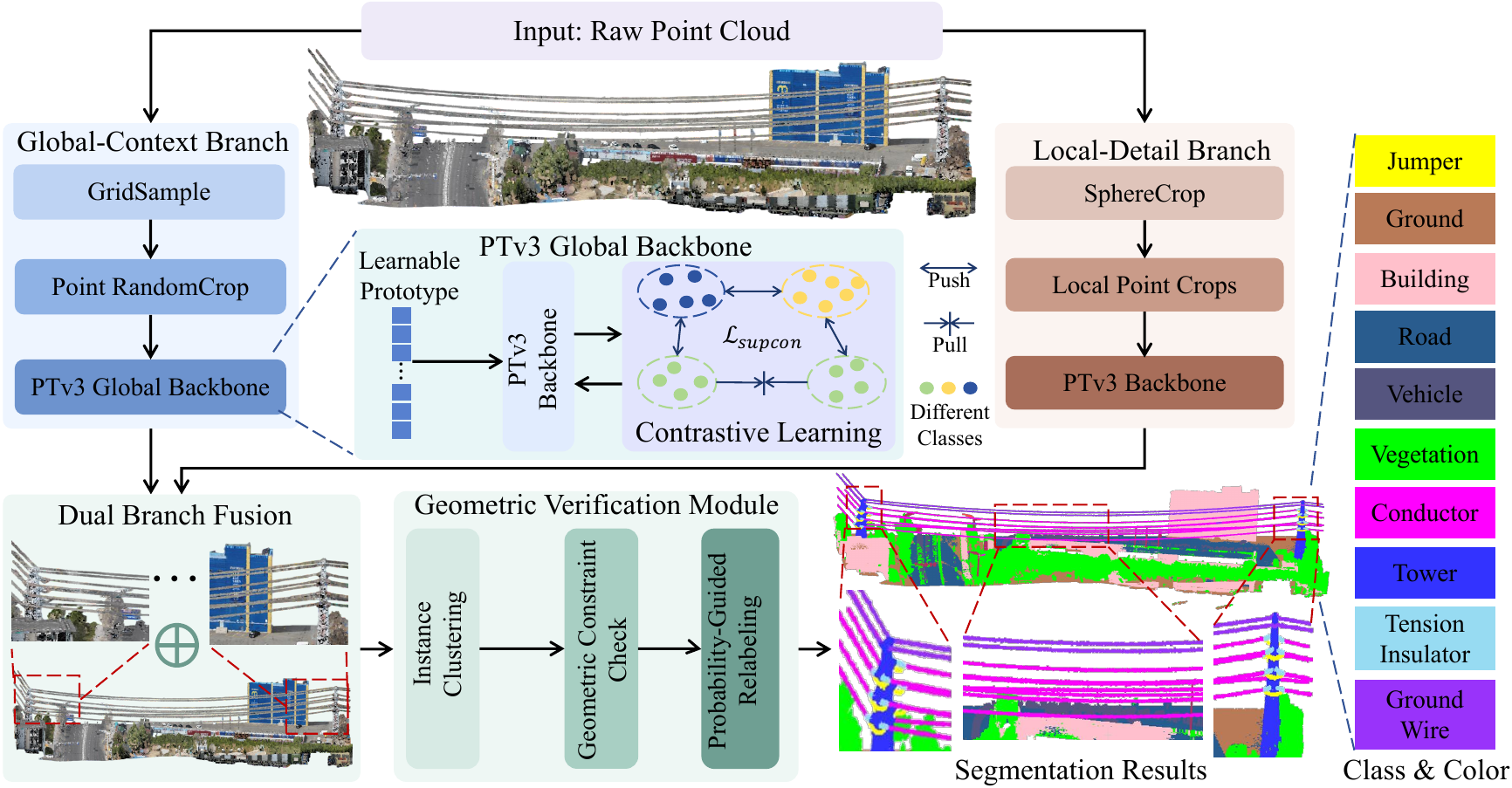}
  \caption{Overview of the proposed global-local fusion framework. A raw transmission-corridor point cloud is processed in parallel by the Global-Context Branch for scene-level structural understanding and the Local-Detail Branch for fine geometric perception. Their prediction probabilities are integrated by Dual Branch Fusion and then corrected by the Geometric Verification Module to produce the final segmentation.}
  \Description{The overall architecture of the proposed framework, including the Global-Context Branch, the Local-Detail Branch, Dual Branch Fusion, and the Geometric Verification Module.}
  \label{fig:pipeline}
\end{figure*}

Fig.~\ref{fig:pipeline} shows the overall framework, which consists of the Global-Context Branch, the Local-Detail Branch, Dual Branch Fusion, and the Geometric Verification Module. The Global-Context Branch processes the raw point cloud to preserve long-range structural context and improve discrimination of confusing categories, while the Local-Detail Branch captures fine geometric details of thin wires and attached components. Their probability outputs are fused and then refined by the Geometric Verification Module. The fused probability for $\mathbf{x}_i$ can be expressed as:

\begin{equation}
p^{\text{fused}}_i = \alpha \, p^{\text{local}}_i + (1-\alpha)\, p^{\text{global}}_i,
\end{equation}
where $p^{\text{local}}_i$ and $p^{\text{global}}_i$ are the probability vectors predicted by the Local-Detail Branch and the Global-Context Branch, respectively, and $\alpha\in[0,1]$ is the fusion weight. The final label is obtained after the fused prediction is processed by the Geometric Verification Module.

\subsection{NoCrop Whole-Scene Training}
\label{subsec:nocrop}

Traditional point cloud segmentation pipelines adopt block-wise partitioning strategies (SphereCrop), which break long-range structural continuity in transmission-corridor scenes. Our NoCrop training strategy preserves global spatial relationships via uniform grid-based downsampling instead of sphere-based cropping. For each point $\mathbf{x}_i$ in the original scene, the downsampled coordinate is computed as:

\begin{equation}
\mathbf{x}'_i = \text{GridSample}(\mathbf{x}_i, s),
\end{equation}
where $s$ is the grid size. The grid-based downsampling divides 3D space into cubic voxels of size $s \times s \times s$ and selects representative points from each voxel, maintaining spatial distribution across the entire scene. To prevent GPU memory overflow while preserving whole-scene representation, we apply RandomCrop with a large point budget:

\begin{equation}
\mathcal{P}' = \text{RandomCrop}(\{\mathbf{x}'_i, {y}'\}_{i=1}^{N'}, N_{\text{max}}),
\end{equation}
where $\{\mathbf{x}'_i, {y}'\}_{i=1}^{N'}$ are the downsampled points with their corresponding labels, $N'$ is the total points, and $N_{\text{max}}$ is the maximum point budget. This allows each iteration to process a substantial scene portion while retaining global context and avoiding memory overflow. We adopt Point Transformer V3 (PTv3) as the backbone for both models for large-scene 3D understanding. For input points $X = \{\mathbf{x}_1, \dots, \mathbf{x}_N\}$, the feature extraction is:

\begin{equation}
F = \Phi_{\text{global}}(X; \Theta_{\text{global}}) = \{\mathbf{f}_1, \dots, \mathbf{f}_N\},
\end{equation}
where $\mathbf{f}_i \in \mathbb{R}^d$ is the feature embedding of point $i$, and $\Phi_{\text{global}}$ denotes the PTv3 Global Backbone in the Global-Context Branch with parameters $\Theta_{\text{global}}$. By processing the whole scene, the Global-Context Branch captures long-range contextual relationships and preserves structural dependencies across large corridor spans.

\subsection{Prototypical Contrastive Learning}
\label{subsec:prototypical}

To enhance discriminability of geometrically similar components and address severe class imbalance, we introduce learnable prototypes  in the Global-Context Branch. First, we apply a projection head to map backbone features to a lower-dimensional embedding space. Unlike a standard classifier, this module introduces extra learnable prototype tokens which provides an additional class-center constraint in a dedicated embedding space.

Second, for each class $c$, we maintain a learnable prototype token $\mathbf{m}_c \in \mathbb{R}^e$ forming the set $\mathbf{M} = \{\mathbf{m}_1, \dots, \mathbf{m}_C\}$, where $e$ is the embedding dimension. These prototype tokens are optimized by back-propagation and act as explicit class centers. The prototype objective pulls each sample toward its true class prototype and pushes it away from other prototypes:

\begin{equation}
\mathcal{L}_{\text{proto}} = -\log \frac{\exp(\mathbf{z}^\top \mathbf{m}_y / \tau)}{\sum_{c=1}^C \exp(\mathbf{z}^\top \mathbf{m}_c / \tau)},
\end{equation}
where $\mathbf{z} \in \mathbb{R}^e$ is the projected embedding, $\mathbf{m}_y$ is the prototype token for the true class $y$, and $\tau = 0.1$ is the temperature parameter.

Beyond prototype-based learning, we incorporate a supervised contrastive loss that directly compares features across samples, encouraging same-class embeddings to cluster together while pushing different-class embeddings apart.

To address extreme class imbalance, we introduce a weighting mechanism for the rare category set $\mathcal{C}_{\text{rare}} = \{0, 12, 13, 14, 18, 19, 20\}$, which includes sparse safety-critical components. For each sample, the weighted prototypical loss is defined as:

\begin{equation}
\mathcal{L}_{\text{proto}}^w = w_y \mathcal{L}_{\text{proto}},
\end{equation}
where $w_y = 5.0$ if $y \in \mathcal{C}_{\text{rare}}$, otherwise $w_y = 1.0$. The overall prototype loss for a mini-batch is then obtained by averaging this weighted loss over all samples.

The overall training objective combines segmentation losses with weighted contrastive terms:

\begin{equation}
\mathcal{L}_{\text{total}} = \mathcal{L}_{\text{CE}} + \mathcal{L}_{\text{Lovasz}} + \lambda_{\text{proto}} \mathcal{L}_{\text{proto}}^w + \lambda_{\text{supcon}} \mathcal{L}_{\text{supcon}}^w,
\end{equation}
where $\mathcal{L}_{\text{CE}}$ and $\mathcal{L}_{\text{Lovasz}}$ denote cross-entropy and mIoU-optimizing Lovasz-Softmax losses, respectively. $\lambda_{\text{proto}} = 0.1$ and $\lambda_{\text{supcon}} = 0.5$ are used to balance the contrastive learning terms with the main segmentation losses. $\mathcal{L}_{\text{supcon}}^w$ is the weighted supervised contrastive loss, which uses the same weighting for rare classes as the prototypical loss.

\subsection{Dual Branch Fusion}
\label{subsec:fusion}

The Global-Context Branch preserves long-range structural context and improves recognition of structure-dependent components, while the Local-Detail Branch retains fine geometric details for thin wires and small attached components. To combine these complementary cues, we adopt Dual Branch Fusion to aggregate the probability predictions from the two branches.

For point $\mathbf{x}_i$, the probability prediction from the Local-Detail Branch is:

\begin{equation}
p^{\text{local}}_i = \Phi_{\text{local}}(\text{SphereCrop}(\mathcal{P}, \mathbf{x}_i, K_{\text{local}}); \Theta_{\text{local}}),
\end{equation}
where $\Phi_{\text{local}}$ denotes the PTv3 Backbone in the Local-Detail Branch with parameters $\Theta_{\text{local}}$, and $K_{\text{local}}$ is the maximum number of points retained in each local crop, set to $120{,}000$ in our implementation. The Local-Detail Branch is more effective for wire-like structures such as conductors and ground wires, where local geometric precision is critical.

The probability prediction from the Global-Context Branch is:

\begin{equation}
p^{\text{global}}_i = \Phi_{\text{global}}(\text{GridSample}(\mathcal{P}, s); \Theta_{\text{global}}),
\end{equation}
where $\Phi_{\text{global}}$ denotes the PTv3 Global Backbone in the Global-Context Branch with parameters $\Theta_{\text{global}}$, and GridSample applies uniform grid downsampling with size $s = 0.25$m. The Global-Context Branch is more effective for context-dependent categories such as jumpers, towers, and tension insulators, where position and attachment relationships are crucial.

Dual Branch Fusion combines the two probability vectors using weighted averaging:

\begin{equation}
p^{\text{fused}}_i = \alpha p^{\text{local}}_i + (1 - \alpha) p^{\text{global}}_i,
\end{equation}
where $p^{\text{local}}_i, p^{\text{global}}_i \in \mathbb{R}^C$ are the probability vectors predicted by the Local-Detail Branch and the Global-Context Branch, respectively, $\alpha \in [0,1]$ is the fusion weight, and $p^{\text{fused}}_i \in \mathbb{R}^C$ is the fused probability vector. A preliminary fused label is obtained by:

\begin{equation}
\tilde{y}_i = \arg\max_{c} p^{\text{fused}}_{i,c},
\end{equation}
then the fused probabilities and preliminary labels are  further refined by the Geometric Verification Module.

\begin{table*}[t]
\centering
\caption{Comparison with other methods on ECLAIR}
\label{tab:eclair_main}
\small
\setlength{\tabcolsep}{3pt}
\begin{tabular*}{\textwidth}{@{\extracolsep{\fill}}lcccccccc@{}}
\toprule
Method & mIoU & T. Wires & D. Wires & Pole & T. Towers & Ground & Veg. & Bldg. \\
\midrule
SpUNet~\cite{pointcept} & 0.6986 & 0.9750 & 0.8296 & 0.7693 & 0.6976 & \textbf{0.9726} & 0.9734 & 0.6268 \\
MinkUNet34C~\cite{minkowski} & 0.6892 & 0.9693 & 0.8286 & 0.7763 & 0.4192 & 0.9723 & \textbf{0.9735} & 0.6107 \\
Stratified Trans.~\cite{stratified} & 0.5484 & 0.9699 & 0.5117 & 0.2670 & 0.2361 & 0.9693 & 0.9693 & 0.6168 \\
PTv2~\cite{ptv2} & 0.5562 & 0.8459 & 0.6789 & 0.0335 & 0.8488 & 0.1128 & 0.9322 & 0.9153 \\
OctFormer~\cite{octformer} & 0.3753 & 0.6012 & 0.0051 & 0.7234 & 0.0589 & 0.0476 & 0.9598 & \textbf{0.9432} \\
PTv3~\cite{ptv3} & 0.7207 & 0.9873 & 0.8656 & 0.7423 & 0.6704 & 0.9701 & 0.9714 & 0.6264 \\
Ours (NoCrop+Proto+Focal) & 0.7825 & 0.9944 & 0.9297 & 0.8010 & 0.8521 & 0.9718 & 0.9730 & 0.7596 \\
Ours (Dual-branch Fusion) & 0.7870 & 0.9950 & 0.9334 & 0.8046 & \textbf{0.8527} & 0.9716 & 0.9728 & 0.7522 \\
Ours (Fusion+Geo) & \textbf{0.7874} & \textbf{0.9963} & \textbf{0.9367} & \textbf{0.8056} & 0.8424 & 0.9716 & 0.9729 & 0.7472 \\
\bottomrule
\end{tabular*}
\end{table*}

\begin{table*}[t]
\centering
\caption{Comparison with other methods on Toronto-3D}
\label{tab:toronto_main}
\small
\setlength{\tabcolsep}{3pt}
\begin{tabular*}{\textwidth}{@{\extracolsep{\fill}}lccccccccc@{}}
\toprule
Method & mIoU & Road & Rd. Mrk & Nat. & Bldg. & U. Line & Pole & Car & Fence \\
\midrule
SpUNet~\cite{pointcept} & 0.6887 & \textbf{0.9462} & 0.0000 & 0.9590 & 0.9135 & 0.8238 & 0.7625 & 0.8617 & 0.2430 \\
MinkUNet34C~\cite{minkowski} & 0.6956 & 0.9448 & 0.0000 & 0.9534 & 0.8953 & 0.8328 & 0.7757 & 0.8964 & 0.2661 \\
Stratified Trans.~\cite{stratified} & 0.5127 & 0.9381 & 0.0000 & 0.8654 & 0.7302 & 0.5762 & 0.2801 & 0.6795 & 0.0317 \\
PTv2~\cite{ptv2} & 0.4426 & 0.8734 & 0.0000 & 0.7539 & 0.6077 & 0.2109 & 0.3870 & 0.5623 & 0.1455 \\
OctFormer~\cite{octformer} & 0.6205 & 0.9250 & 0.0000 & 0.9259 & 0.8189 & 0.8099 & 0.6698 & 0.6285 & 0.1863 \\
PTv3~\cite{ptv3} & 0.7086 & 0.9458 & 0.0000 & 0.9538 & \textbf{0.9273} & 0.8619 & 0.7983 & 0.8973 & 0.2846 \\
Ours (NoCrop+Proto+Focal) & 0.7253 & 0.9433 & 0.0000 & 0.9541 & 0.9099 & 0.8661 & 0.7910 & 0.8826 & 0.4555 \\
Ours (Dual-branch Fusion) & 0.7251 & 0.9459 & 0.0000 & 0.9580 & 0.9228 & 0.8677 & 0.8191 & 0.9134 & 0.3741 \\
Ours (Fusion+Geo) & \textbf{0.7371} & 0.9459 & 0.0000 & \textbf{0.9600} & 0.9185 & \textbf{0.8682} & \textbf{0.8224} & \textbf{0.9227} & \textbf{0.4588} \\
\bottomrule
\end{tabular*}
\end{table*}

\subsection{Geometric Verification Module}
\label{subsec:geometric}

Although Dual Branch Fusion improves the accuracy of predictions, the preliminary fused labels for small and geometrically similar components may still violate basic physical constraints. The Geometric Verification Module refines these results with class-specific geometric rules. This module first groups preliminarily labeled points into candidate component instances using DBSCAN, and then performs constraint-based verification followed by probability-guided relabeling.

For class $c$, we collect the points preliminarily assigned to that class as $X_c = \{\mathbf{x}_i \mid \tilde{y}_i = c\}$. Then DBSCAN will be used to extract candidate instances:

\begin{equation}
\mathcal{I}_c = \text{DBSCAN}(X_c, \epsilon_c, \text{min\_samples}_c) = \{I_1, I_2, \dots, I_K\},
\end{equation}
where $\mathcal{I}_c$ is the set of candidate instances for class $c$, $I_j$ denotes the $j$-th cluster in this set, $K$ is the total number of clusters detected for class $c$, and $\epsilon_c$ and $\text{min\_samples}_c$ are class-specific clustering parameters. DBSCAN automatically discovers arbitrary-shaped clusters without requiring prior knowledge of the number of instances, which is suitable for the varying component counts in different tower structures.

For each cluster $I_j$, geometric verification evaluates whether the predicted instance is physically valid. The verification score is defined as:

\begin{equation}
\text{Score}(I_j) = \frac{1}{|\mathcal{S}(I_j)|} \sum_{s \in \mathcal{S}(I_j)} w_s \cdot \mathbb{I}(\text{constraint}_s \text{ is satisfied}),
\end{equation}
where $\text{Score}(I_j)$ measures the geometric plausibility of cluster $I_j$, $\mathcal{S}(I_j)$ denotes the set of applicable constraints for that cluster, $w_s$ are normalized constraint weights, and $\mathbb{I}(\cdot)$ is the indicator function that equals $1$ when the corresponding constraint is satisfied and $0$ otherwise. For tension insulators, we compute the orientation ratio
\begin{equation}
\text{Orientation}(I_j) = \frac{|(\mathbf{x}_{\text{end}} - \mathbf{x}_{\text{start}})_z|}{\|\mathbf{x}_{\text{end}} - \mathbf{x}_{\text{start}}\|_2},
\end{equation}
where $\mathbf{x}_{\text{start}}$ and $\mathbf{x}_{\text{end}}$ are the two endpoints of the cluster. Valid tension insulators should have a small vertical ratio relative to their total length, while line insulators are expected to satisfy $\text{Orientation}(I_j) \approx 1$.

When the verification score falls below the threshold $\tau_{\text{geo}} = 0.4$, the points in the corresponding cluster are relabeled by:

\begin{equation}
\hat{y}_i = \begin{cases}
\arg\max_{c \neq \tilde{y}_i} p^{\text{fused}}_{i,c}, & \text{if } \mathbf{x}_i \in I_j \text{ and } \text{Score}(I_j) < \tau_{\text{geo}} \\
\tilde{y}_i, & \text{otherwise}
\end{cases},
\end{equation}
where $\hat{y}_i$ is the final label after geometric verification. We build a KD-Tree to accelerate neighborhood queries during verification. This module mainly removes structurally implausible predictions and improves discrimination of small and confusing components.

\section{Experiments}
\label{sec:exp}

\paragraph{Implementation and evaluation protocol}
We use the official splits for all datasets. All models are implemented in Pointcept and trained for 100 epochs with AdamW (base learning rate $3\times10^{-3}$ and weight decay $0.05$) and OneCycle scheduling. On TowerDataset, the global branch uses 0.25 m NoCrop whole-scene training with batch size 1, a random crop of up to 4M voxelized points, and the Proto+Focal setting. During inference, we apply standard test-mode voxelization and inverse mapping, fuse local and global predictions by weighted probability averaging with validation-tuned $\alpha$ (0.50 on TowerDataset, 0.30 on ECLAIR, and 0.66 on Toronto-3D), and then perform geometric validation for attachment classes. On TowerDataset, we report 22-class mIoU on the 120-scene test split together with representative class IoUs. On ECLAIR~\cite{eclair} and Toronto-3D~\cite{toronto3d}, we follow the same optimization recipe with voxel sizes of 0.20 m and 0.06 m, and report the official mIoU and class-wise IoU on the standard evaluation splits. All experiments are conducted on a single NVIDIA RTX A6000 Pro GPU with 96 GB memory.

\paragraph{Compared methods}
\textbf{SpUNet~\cite{pointcept}} is a sparse U-Net baseline with efficient encoder-decoder design. \textbf{MinkUNet34C~\cite{minkowski}} uses voxel-based sparse convolutions as a strong baseline. \textbf{Stratified Transformer~\cite{stratified}} introduces stratified window attention for dense point clouds. \textbf{PTv2~\cite{ptv2}} improves point transformer with grouped vector attention. \textbf{OctFormer~\cite{octformer}} adopts octree-based transformer for large scenes. \textbf{PTv3~\cite{ptv3}} simplifies transformer pipeline with strong scalability. Together, these baselines cover sparse convolution, point transformer, window attention, and octree paradigms, providing a representative and diverse benchmark for large-scale outdoor point cloud segmentation.

\subsection{Comparison with State-of-the-Art Methods}
We report quantitative comparison on three datasets: ECLAIR for same-domain evaluation, Toronto-3D for cross-domain evaluation, and TowerDataset as the main benchmark.

\begin{table*}[t]
\centering
\caption{Comparison with state-of-the-art methods on TowerDataset under the deduplicated 120-scene test split}
\label{tab:tower_main}
\scriptsize
\resizebox{\textwidth}{!}{%
\begin{tabular}{lcccccccccc}
\toprule
Method & 22-class mIoU & Jumper & Str. Ins. & V-Ins. & Lin. Ins. & Spacer & Opt. Cab. & Cond. & G. Wire & Tower \\
\midrule
SpUNet~\cite{pointcept} & 0.3280 & 0.8155 & 0.7094 & 0.6838 & 0.5146 & 0.0000 & 0.0000 & 0.9469 & \textbf{0.9466} & 0.8953 \\
MinkUNet34C~\cite{minkowski} & 0.3362 & 0.8599 & 0.7503 & \textbf{0.7704} & 0.5211 & 0.0000 & 0.0000 & \textbf{0.9472} & 0.9449 & 0.9060 \\
Stratified Trans.~\cite{stratified} & 0.2874 & 0.0115 & 0.4637 & 0.1463 & 0.4902 & 0.3146 & 0.1152 & 0.8592 & 0.7865 & 0.7761 \\
PTv2~\cite{ptv2} & 0.3553 & \textbf{0.8997} & \textbf{0.8107} & 0.7269 & 0.5665 & 0.0000 & 0.0000 & 0.9304 & 0.9073 & \textbf{0.9451} \\
OctFormer~\cite{octformer} & 0.1794 & 0.3518 & 0.4276 & 0.3147 & 0.3618 & 0.0000 & 0.0000 & 0.4795 & 0.2121 & 0.5023 \\
PTv3~\cite{ptv3} & 0.4116 & 0.8859 & 0.7131 & 0.3276 & 0.6056 & \textbf{0.8062} & 0.2327 & 0.8884 & 0.7952 & 0.8976 \\
Ours & \textbf{0.4369} & 0.8844 & 0.7522 & 0.4455 & \textbf{0.6430} & 0.8059 & \textbf{0.2635} & 0.9154 & 0.8362 & 0.9156 \\
\bottomrule
\end{tabular}}
\end{table*}

\begin{table*}[t]
\centering
\caption{Ablation experiments on TowerDataset}
\label{tab:ablation_modules}
\small
\setlength{\tabcolsep}{3pt}
\resizebox{\textwidth}{!}{%
\begin{tabular}{ccccccccccccccccc}
\toprule
NoCrop & Proto & Focal & Fusion & GeoValid & Primary mIoU & 22-mIoU & Jumper & Str. Ins. & V-Ins. & Lin. Ins. & Spacer & Opt. Cab. & Cond. & G. Wire & Tower & Low Veg. \\
\midrule
-- & -- & -- & -- & -- & 0.7018 & 0.4116 & 0.8859 & 0.7131 & 0.3276 & 0.6056 & 0.8062 & 0.2327 & 0.8884 & 0.7952 & 0.8976 & 0.8653 \\
\checkmark & -- & -- & -- & -- & 0.7025 & 0.4084 & 0.9023 & 0.7183 & \textbf{0.4702} & 0.6342 & 0.7690 & 0.2537 & 0.8345 & 0.7998 & 0.8041 & 0.8385 \\
\checkmark & \checkmark & -- & -- & -- & 0.6795 & 0.4148 & \textbf{0.9212} & 0.7338 & 0.4533 & 0.6355 & 0.8004 & \textbf{0.2811} & 0.8978 & 0.7668 & 0.4873 & 0.8180 \\
\checkmark & \checkmark & \checkmark & -- & -- & 0.7114 & 0.4220 & 0.8584 & 0.7387 & 0.4307 & 0.6274 & 0.7528 & 0.2310 & 0.9154 & 0.8260 & 0.8989 & 0.8349 \\
\checkmark & \checkmark & -- & \checkmark & -- & 0.7194 & 0.4336 & 0.9166 & 0.7446 & 0.4684 & 0.6384 & \textbf{0.8142} & 0.2667 & 0.9102 & 0.8084 & 0.7655 & 0.8613 \\
\checkmark & \checkmark & \checkmark & \checkmark & -- & 0.7323 & 0.4365 & 0.8844 & 0.7522 & 0.4396 & 0.6430 & 0.8059 & 0.2635 & 0.9154 & 0.8362 & 0.9156 & 0.8669 \\
\checkmark & \checkmark & \checkmark & \checkmark & \checkmark & \textbf{0.7336} & \textbf{0.4369} & 0.8844 & \textbf{0.7522} & 0.4455 & \textbf{0.6430} & 0.8059 & 0.2635 & \textbf{0.9154} & \textbf{0.8362} & \textbf{0.9156} & \textbf{0.8669} \\
\bottomrule
\end{tabular}}
\end{table*}

\paragraph{Results on ECLAIR}
ECLAIR is the public benchmark related to our target application. As shown in Table~\ref{tab:eclair_main}, the proposed full model achieves the best performance with 0.7874 mIoU, improving over PTv3 from 0.7207 by 6.67 points. The NoCrop+Proto+Focal variant already reaches 0.7825, which shows that the Global-Context Branch and Learnable Prototype Contrastive Learning provide a strong single-model baseline on this dataset. After introducing Dual Branch Fusion, the mIoU further rises to 0.7870, and the Geometric Verification Module brings the final result to 0.7874.

The gains mainly come from the core power-corridor categories. Compared with PTv3, our full method improves distribution wires from 0.8656 to 0.9367, and poles from 0.7423 to 0.8056. These improvements indicate that the proposed framework can better preserve wire continuity and reduce confusion around some vertical structures. Tower recognition also benefits from the proposed design. The Dual Branch Fusion model reaches the best tower IoU of 0.8527, far above the 0.6704 of PTv3, which confirms that combining global structure and local detail is especially effective for large structural components. After geometric verification, tower IoU remains high at 0.8424 while the overall mIoU becomes the best among all methods, showing that the final verification step improves global prediction quality without sacrificing the main corridor categories.

\paragraph{Results on Toronto-3D}
This dataset evaluates cross-domain robustness from aerial scanning to mobile LiDAR. As shown in Table~\ref{tab:toronto_main}, our full model achieves the best result with 0.7371 mIoU, improving over PTv3 from 0.7086 by 2.85 points. The NoCrop+Proto+Focal variant already reaches 0.7253, which indicates that the Global-Context Branch and Learnable Prototype Contrastive Learning generalize well under domain shift. Dual Branch Fusion gives a comparable overall result of 0.7251, but it improves several important categories, including road, natural terrain, building, utility line, pole, and car. After that, the Geometric Verification Module raises the final mIoU to 0.7371.

The final gains are on sparse and difficult categories. Compared with PTv3, our full model improves fence from 0.2846 to 0.4588, car from 0.8973 to 0.9227 and utility line from 0.8619 to 0.8682. These results show that the proposed framework remains effective when the data distribution shifts from aerial scenes to mobile scanning scenes. Building accuracy stays high at 0.9185, although the best single value on this category is 0.9273 from PTv3. Road markings remain 0.0000 for all methods because this class is extremely sparse in the official test split.

\paragraph{Results on TowerDataset}
TowerDataset serves as our primary benchmark because it contains fine-grained component categories and severe class imbalance. As shown in Table~\ref{tab:tower_main}, our method achieves the best result with 0.4369 in 22-class mIoU. The main gains come from ambiguous attachment categories. Compared with PTv3, our method improves strain insulator from 0.7131 to 0.7522, V-string insulator from 0.3276 to 0.4455, line insulator from 0.6056 to 0.6430 and optical cable from 0.2327 to 0.2635. These improvements show that combining global structure and local detail is effective for categories that are locally similar but differ in attachment pattern and corridor context.

Table~\ref{tab:tower_main} also shows that existing methods still struggle to achieve balanced performance on this benchmark. SpUNet, MinkUNet34C, PTv2, and OctFormer all obtain zero IoU on spacer and optical cable, which indicates that conventional large-scene backbones remain weak under the extreme long-tail setting. PTv2 reaches the best scores on jumper, strain insulator, and tower, and MinkUNet34C reaches the best V-string insulator and conductor scores, but these methods do not maintain strong performance across the full set of rare and dominant categories. In contrast, our method delivers the highest overall 22-class mIoU while keeping strong performance on the main power-line classes and improving the difficult attachment categories, which is consistent with the design goal of Dual Branch Fusion and the Geometric Verification Module.

\begin{figure*}[t]
\centering
\includegraphics[width=1\textwidth]{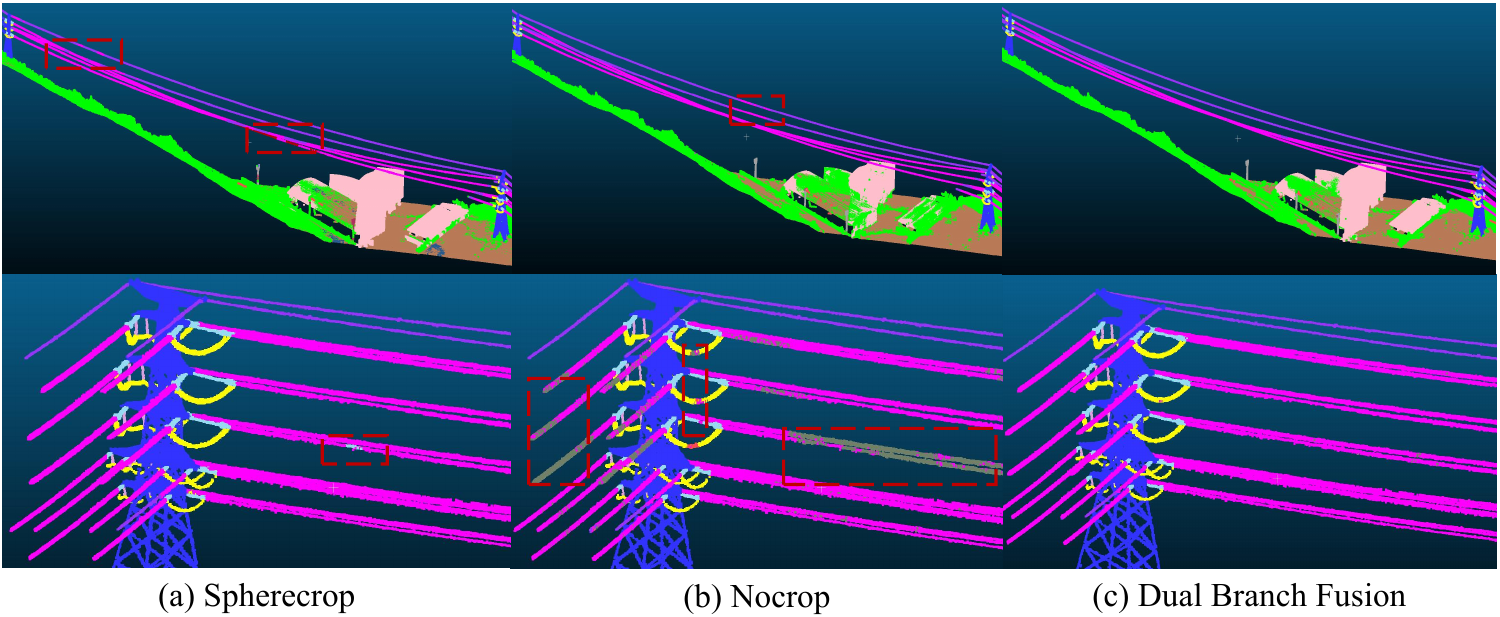}
\caption{Qualitative comparison of different prediction strategies on TowerDataset. The first row shows overall scene-level results, and the second row shows enlarged local details. From left to right, the three columns correspond to SphereCrop, NoCrop, and Fusion.}
\Description{A qualitative comparison figure on TowerDataset with two rows. The first row shows overall predictions for representative scenes, and the second row shows enlarged detail regions. The three columns correspond to SphereCrop, NoCrop, and Fusion.}
\label{fig:vis_compare}
\end{figure*}

\subsection{Ablation Studies}
Ablation experiments on TowerDataset are summarized in Table~\ref{tab:ablation_modules}. We report primary 10-class mIoU, 22-class mIoU, and IoUs for the ten primary categories.

NoCrop improves attachment recognition by preserving long-range topology, like raising V-string insulator from 0.3276 to 0.4702, but it also weakens dominant structural classes, with tower dropping from 0.8976 to 0.8041. Learnable Prototype Contrastive Learning further enhances rare-class discrimination and increases 22-class mIoU to 0.4148, while Focal reweighting restores the balance between rare and dominant classes and gives the strongest single-model result of 0.4220. On this basis, Dual Branch Fusion and the Geometric Verification Module provide further gains, yielding the best results of 0.7336 primary 10-class mIoU and 0.4369 22-class mIoU. These results demonstrate that each module contributes to the final model, and that robust transmission-corridor segmentation benefits from jointly modeling global context, local detail, and class-balanced optimization.

\subsection{Efficiency and Computational Cost}

\begin{table}[t]
\centering
\caption{Efficiency and computational cost comparison on TowerDataset}
\label{tab:efficiency_cost}
\small
\setlength{\tabcolsep}{2pt}
\begin{tabular*}{\columnwidth}{@{\extracolsep{\fill}}p{0.42\columnwidth}ccc@{}}
\toprule
Method & AvgFwd (s) & AvgPipe (s) & PeakAlloc (GB) \\
\midrule
SpUNet~\cite{pointcept} & 28.0 & 82.7 & 2.3 \\
MinkUNet34C~\cite{minkowski} & 82.0 & 168.8 & 2.9 \\
PTv2~\cite{ptv2} & 809.9 & 877.1 & 24.2 \\
OctFormer~\cite{octformer} & 795.6 & 871.5 & 24.0 \\
PTv3~\cite{ptv3} & 171.0 & 467.2 & 0.6 \\
Ours (NoCrop) & 47.0 & 102.8 & 3.0 \\
Ours (Dual-Branch Fusion) & 206.4 & 581.4 & 3.1 \\
\bottomrule
\end{tabular*}
\end{table}

Table~\ref{tab:efficiency_cost} compares the computational cost of representative baselines and our methods on TowerDataset. SpUNet is the fastest baseline in both forward and end-to-end pipeline time, while PTv3 has the lowest peak memory allocation at only $0.6$ GB. PTv2 and OctFormer are the most expensive methods, each requiring about $24$ GB peak memory and around $800$ seconds average forward time. Compared with PTv3, NoCrop is much faster in both forward and pipeline time at the cost of higher memory, while Dual-Branch Fusion further increases runtime but still uses far less memory than PTv2 and OctFormer.

\subsection{Visualization Result Comparison}

Fig.~\ref{fig:vis_compare} gives a qualitative comparison of the three strategies on TowerDataset. The first row shows  scene-level results. SphereCrop preserves part of the local geometry while breaking the continuity of long wires and attached components. NoCrop restores the global corridor topology and produces more coherent predictions. However, relying only on whole-scene reasoning still leaves some ambiguous regions around thin components. Fusion combines the strengths of the two branches with better continuity along wires and fewer isolated errors around attachment regions.
The second row shows enlarged local regions where SphereCrop often produces fragmented predictions and misclassification around small components, while NoCrop improves continuity but still leaves some coarse boundaries and category confusion. Fusion yields cleaner shapes and more stable predictions for insulators and nearby components, with fewer scattered errors around the supporting structures. These visual results are consistent with the quantitative results and show that the proposed framework better balances global consistency and local detail.

\section{Conclusion}
\label{sec:conclusion}
In this study, we introduced TowerDataset to address the lack of realistic benchmarks for fine-grained transmission-corridor segmentation, as existing datasets often provide only coarse labels or fragmented scenes and cannot adequately evaluate long-range structure or rare component recognition. Building on this benchmark, we proposed a global-local fusion framework integrating NoCrop whole-scene learning, Learnable Prototype Contrastive Learning, Dual Branch Fusion, and the Geometric Verification Module to better preserve corridor topology and improve discrimination of sparse and ambiguous components. Extensive experiments on TowerDataset and public benchmarks demonstrate robust and consistent performance and ablation studies also confirm the effectiveness of each component.

Several limitations remain for future work. NoCrop training increases memory requirements, although large-point RandomCrop partly mitigates this issue. Dual Branch Fusion still relies on manual weight tuning, and adaptive uncertainty-aware strategies may improve performance further. The Geometric Verification Module assumes known attachment relationships, so learning such constraints from data may enhance generalization. In addition, multimodal fusion with optical imagery could further improve discrimination of geometrically similar components.

\bibliographystyle{ACM-Reference-Format}
\bibliography{references}

\clearpage
\appendix
\section{Additional Dataset Materials}
\label{sec:appendix_dataset}

\subsection{Dataset Overview}
This subsection briefly summarizes the released TowerDataset. The current release comprises $661$ preprocessed scenes, with a total of $2.466$ billion points and a storage footprint of approximately $78.73$ GiB. Each scene is stored as a \texttt{.pth} dictionary containing three fields: \texttt{coord}, \texttt{color}, and \texttt{segment}, where $N$ denotes the number of points in the scene. Specifically, \texttt{coord} is an $N\times3$ matrix recording the $(x,y,z)$ location of each point, \texttt{color} is an $N\times3$ matrix containing per-point RGB attributes, and \texttt{segment} is an $N$-dimensional vector storing the semantic label of each point. This representation preserves geometric, appearance, and annotation information at the point level. In the experiments reported in this paper, however, the model input is restricted to \texttt{coord} during both training and inference; neither \texttt{color} nor any other auxiliary attribute is used as input. The original semantic annotations retain $22$ class IDs. Under the primary evaluation protocol, high vegetation is merged into low vegetation, yielding a $10$-class power-corridor subset.

As shown in Table~\ref{tab:appendix_split_stats}, we report the number of scenes, total points, average and median point counts, median projected density, and median major-axis corridor length for each split and for the full dataset. These statistics show that TowerDataset is not concentrated around one narrow scene pattern: the training split covers the broadest source range, whereas the validation and test splits are denser and slightly larger on average. This confirms that the released benchmark portion contains structurally nontrivial corridors rather than only routine short spans.

\begin{table}[t]
\centering
\caption{Split statistics of TowerDataset, reporting scene count, total points, typical per-scene scale, projected density, and corridor major-axis length for each released split}
\label{tab:appendix_split_stats}
\small
\setlength{\tabcolsep}{3pt}
\begin{tabular*}{\columnwidth}{@{\extracolsep{\fill}}lcccccc@{}}
\toprule
Split & Scenes & Points & Avg Mpts & Med. Mpts & Med. Den. & Med. Len. \\
\midrule
Train & 419 & 1.347B & 3.21 & 2.04 & 73.9 & 357.1 \\
Val & 122 & 0.556B & 4.56 & 3.25 & 134.4 & 369.7 \\
Test & 120 & 0.563B & 4.69 & 2.61 & 139.1 & 339.9 \\
Total & 661 & 2.466B & 3.73 & 2.33 & 93.8 & 355.8 \\
\bottomrule
\end{tabular*}
\end{table}

\subsection{Dataset Statistics}
As shown in Figure~\ref{fig:appendix_scene_stats}, we report the per-scene distributions of point count, major-axis corridor length, and projected density. The median scene contains $2.33$M points and spans $355.8$ m along the major axis, while the $90$th percentile reaches $7.76$M points and $641.5$ m. These statistics show that TowerDataset consists of long-span corridor scenes with substantial scale variation. Compared with the training split, the validation and test splits also contain more points on average and exhibit higher projected density.

Here, projected density is computed as the point count divided by the footprint of the XY oriented bounding box. In addition to the geometric variation shown in Figure~\ref{fig:appendix_scene_stats}, the dataset also covers multiple engineering settings, including $110$kV, $220$kV, and $500$kV corridors, as well as several tower structures.

\begin{figure*}[t]
\centering
\includegraphics[width=\textwidth]{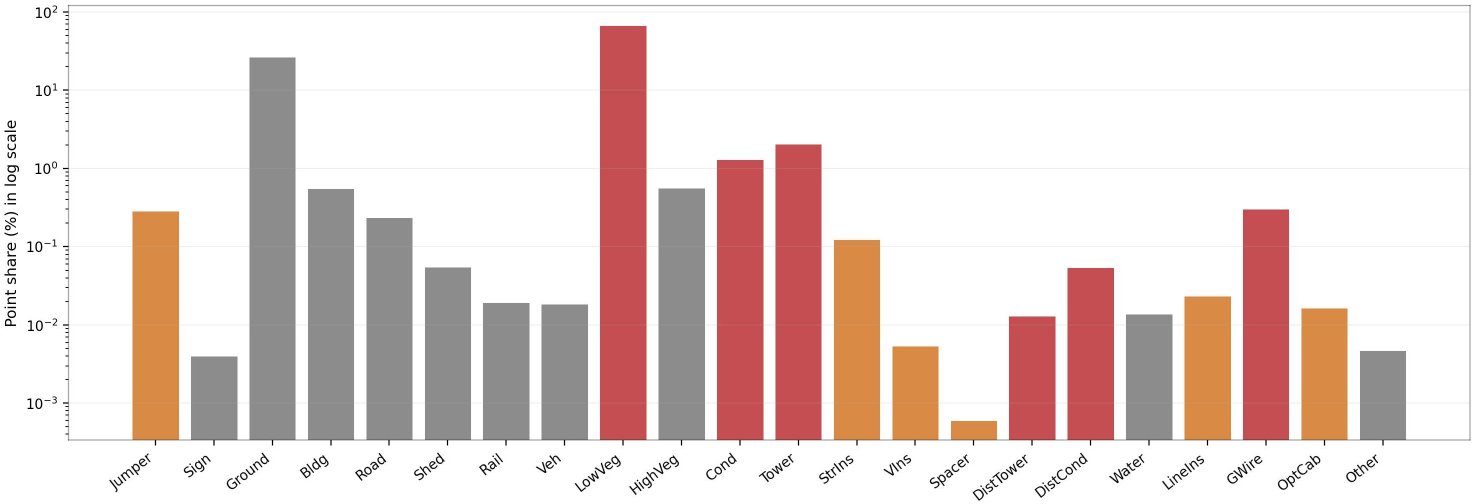}
\caption{Raw $22$-class point distribution of TowerDataset in log scale. Primary power-line classes are highlighted in red, the thinnest critical parts are shown in orange, and other context classes are shown in gray. The logarithmic scale emphasizes that many engineering-critical components occupy several orders of magnitude fewer points than ground and vegetation.}
\Description{A log-scale bar chart of the raw 22-class point distribution of TowerDataset. Primary corridor classes are colored red, critical thin parts are orange, and other context classes are gray, highlighting the extreme point-count imbalance between sparse attachments and dominant background categories.}
\label{fig:appendix_class_tail}
\end{figure*}

As shown in Figure~\ref{fig:appendix_class_tail} and Table~\ref{tab:appendix_semantic_groups}, we further summarize the full $22$-class annotations by merged semantic groups. Here, \textit{transmission backbone} includes conductor, tower, and ground wire; \textit{critical attachments} includes jumper, strain insulator, V-string insulator, line insulator, and spacer; and \textit{distribution assets} includes distribution tower and distribution conductor. Ground and vegetation account for $94.94\%$ of all points, whereas critical attachments, optical cable, and distribution assets account for only $0.43\%$, $0.0165\%$, and $0.0677\%$, respectively. This distribution indicates a pronounced long-tail property in the semantic composition of TowerDataset.

\begin{table}[t]
\centering
\caption{Semantic group statistics of TowerDataset, highlighting the strong imbalance between dominant context classes and geometrically critical but extremely sparse attachment classes}
\label{tab:appendix_semantic_groups}
\small
\setlength{\tabcolsep}{4pt}
\begin{tabular*}{\columnwidth}{@{\extracolsep{\fill}}lrr@{}}
\toprule
Group & Points & Share \\
\midrule
Ground + vegetation & 2,341,392,941 & 94.94\% \\
Transmission backbone & 89,685,190 & 3.64\% \\
Critical attachments & 10,722,737 & 0.43\% \\
Optical cable & 407,159 & 0.0165\% \\
Distribution assets & 1,669,453 & 0.0677\% \\
Other context & 22,199,507 & 0.90\% \\
\bottomrule
\end{tabular*}
\end{table}

\begin{table*}[t]
\centering
\caption{Scene occurrence of primary classes and their typical appearance contexts in TowerDataset, showing which categories are common scene-wide and which remain rare even at the scene level}
\label{tab:appendix_diversity}
\small
\setlength{\tabcolsep}{4pt}
\begin{tabular}{lcc p{0.58\textwidth}}
\toprule
Primary class & Scenes & Ratio (\%) & Typical appearance contexts \\
\midrule
Conductor & 660 & 99.8 & Nearly all corridor spans, including routine 110/220/500 kV transmission-line scenes \\
Tower & 630 & 95.3 & Scenes containing one or two support structures together with tower--wire transitions \\
Ground wire & 652 & 98.6 & Main transmission spans with shield wires above the conductor bundle \\
Low vegetation & 648 & 98.0 & Terrain and vegetation background surrounding most corridor scenes \\
Jumper & 429 & 64.9 & Tension spans and attachment-rich tower-end regions \\
Strain insulator & 394 & 59.6 & Tension towers, dead-end attachments, and line-turn regions \\
Line insulator & 611 & 92.4 & Routine suspension spans near towers and cross-arm attachment locations \\
V-string insulator & 106 & 16.0 & Dedicated V-string or UHV attachment scenes with rarer suspension layouts \\
Optical cable & 104 & 15.7 & Corridors containing OPGW or other communication-cable attachments \\
Spacer & 44 & 6.7 & Long bundled-conductor spans where spacer fittings are explicitly visible \\
\bottomrule
\end{tabular}
\end{table*}

\subsection{Primary-Class Occurrence}
As shown in Table~\ref{tab:appendix_diversity}, we report the number of scenes and occurrence ratio of each primary class, together with their typical appearance contexts. Conductor, ground wire, tower, and low vegetation appear in most scenes, whereas several thin attachment classes remain much less frequent. In particular, V-string insulators, optical cables, and spacers are observed in only $106$ ($16.0\%$), $104$ ($15.7\%$), and $44$ ($6.7\%$) scenes, respectively.

These scene-level statistics complement the point-level distribution and further indicate that the rare attachment classes are sparse not only in total points but also in scene occurrence.

\begin{figure*}[!t]
\centering
\includegraphics[width=\textwidth]{appendix_scene_stats.pdf}
\caption{Distribution of per-scene point count, corridor major-axis length, and projected density across the $661$-scene TowerDataset. The orange and green dashed lines indicate the median and $90$th percentile, respectively, making the heavy-tailed variation in scene scale and sampling density explicit rather than relying only on average values.}
\Description{Three histograms showing points per scene in millions, corridor major-axis length in meters, and projected density in points per square meter for the 661-scene TowerDataset release. Each plot includes median and 90th-percentile markers, illustrating the strong long-tail variation in corridor scale and point sampling density.}
\label{fig:appendix_scene_stats}
\end{figure*}

\begin{figure*}[t]
\centering
\includegraphics[width=1\textwidth]{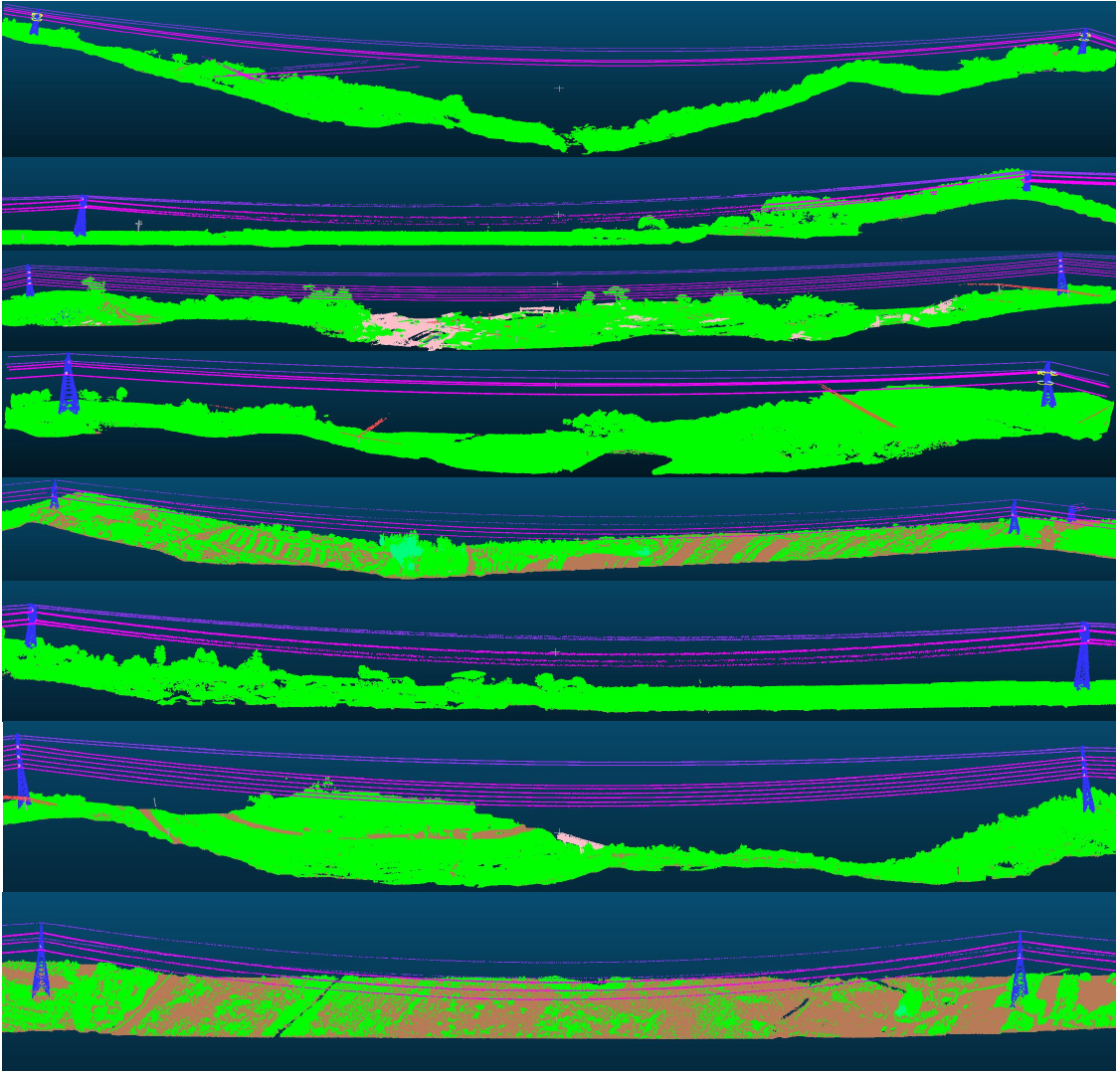}
\caption{Representative corridor scenes from TowerDataset. The displayed panels are drawn from an eight-scene set consisting of Fengshui 29--30, Zhujiang 0003, Beimu 2--3, Fengshui 40--41, Xiaguang 63--64, Zhujiang 0002, Beimu 3--4, and Xiaguang 75--76.}
\Description{Eight TowerDataset corridor scenes drawn from an eight-scene set including Fengshui 29--30, Zhujiang 0003, Beimu 2--3, Fengshui 40--41, Xiaguang 63--64, Zhujiang 0002, Beimu 3--4, and Xiaguang 75--76, covering multiple voltage levels, span lengths, point counts, clutter levels, and support configurations.}
\label{fig:appendix_scene_gallery_main}
\end{figure*}

\begin{figure*}[t]
\centering
\includegraphics[width=1\textwidth]{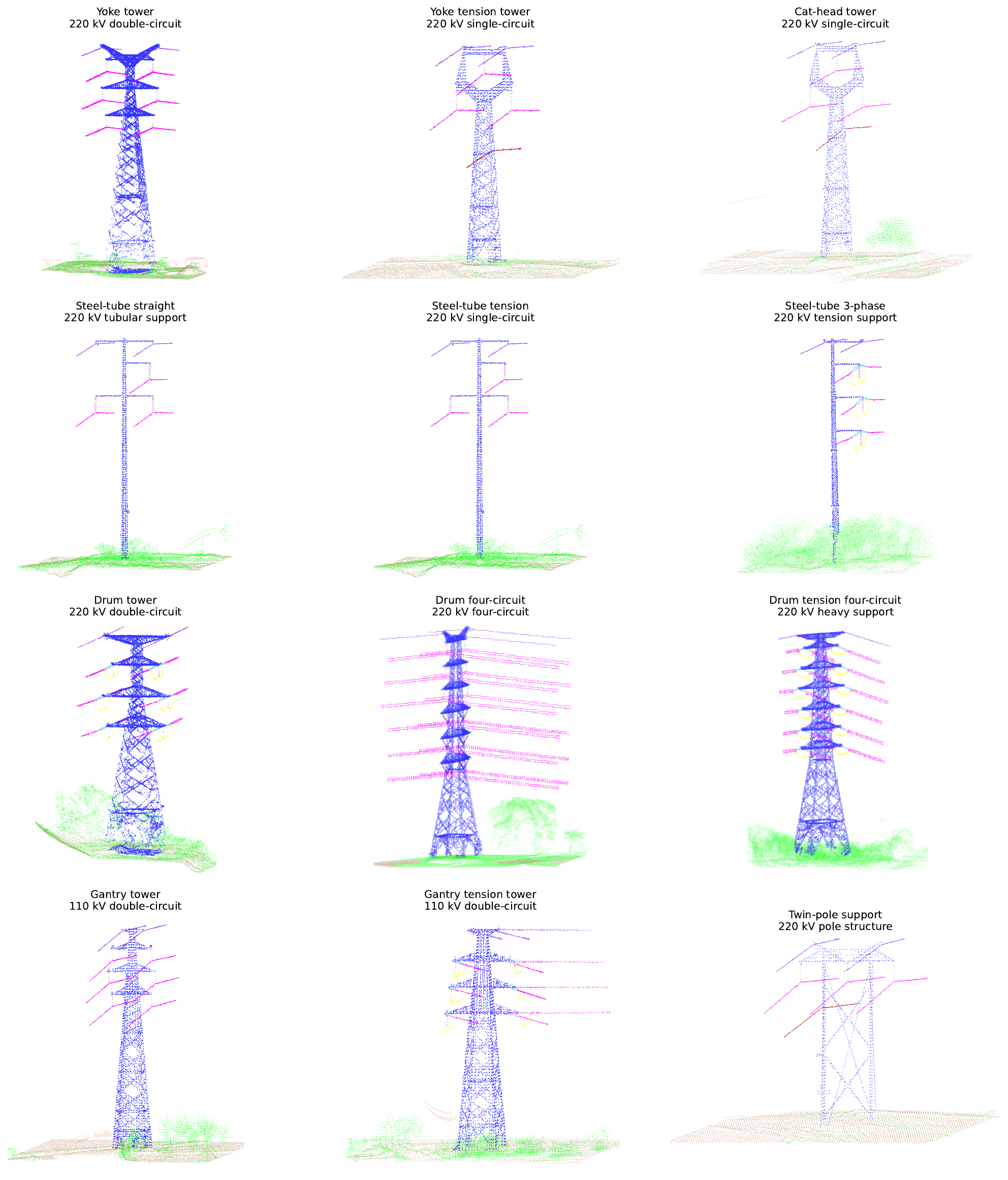}
\caption{Twelve representative tower and pole types in TowerDataset.}
\Description{A 4 by 3 gallery of twelve representative tower and pole types in TowerDataset, including multiple support families and several straight-versus-tension variants rendered from annotated point clouds.}
\label{fig:appendix_tower_type_gallery}
\end{figure*}

\subsection{Representative Scenes} 
As shown in Fig.~\ref{fig:appendix_scene_gallery_main}, this subsection presents scene-level visualizations of eight representative large-scale scenes from TowerDataset: Fengshui 29--30, Zhujiang 0003, Beimu 2--3, Fengshui 40--41, Xiaguang 63--64, Zhujiang 0002, Beimu 3--4, and Xiaguang 75--76. Together, these scenes cover 110\,kV, 220\,kV, and 500\,kV power-line settings, with corridor lengths ranging from 546.9\,m to 971.7\,m and point counts from 1.13M to 9.12M. These examples illustrate variation in corridor extent, sampling density attachment concentration, and others.

Among them, Fengshui 29--30, Zhujiang 0003, Zhujiang 0002, and Fengshui 40--41 represent large 220\,kV corridors, with spans of 971.7\,m, 810.2\,m, 699.2\,m, and 546.9\,m and point counts of 7.70M, 8.39M, 6.65M, and 7.95M, respectively. Beimu 2--3 and Beimu 3--4 are denser 110\,kV corridors with 9.12M and 4.11M points over 602.6\,m and 566.5\,m, while Xiaguang 63--64 and Xiaguang 75--76 show two 500\,kV V-string configurations with 1.51M and 1.13M points over 904.9\,m and 876.8\,m.

Beyond scene-scale variation, TowerDataset also contains diverse tower and pole types. Figure~\ref{fig:appendix_tower_type_gallery} shows $12$ representative examples from the original scenes, including yoke-type, cat-head, steel-tube, drum-type, gantry-type, and twin-pole structures, as well as straight and tension variants in several families. These examples show differences in structure and appearance, confirming the diversity of tower and pole types in TowerDataset.
Such diversity makes TowerDataset more challenging. Specifically, scene spans and point densities vary across corridors, while thin attachments constitute only a tiny fraction of points. Additionally, tower and pole structures evolve with voltage levels and line configurations. Models therefore need to capture long-range corridor context, preserve tiny structural components, and maintain robustness against vegetation clutter and structural heterogeneity.

\end{document}